\documentclass{article}
\pdfoutput=1

\usepackage[utf8]{inputenc}
\usepackage[T1]{fontenc}    % use 8-bit T1 fonts
\usepackage{graphicx}
\usepackage[numbers]{natbib}
\usepackage{booktabs}
\usepackage{tabularx}
\usepackage{ragged2e}
\usepackage{xcolor}
\usepackage{subcaption}
\usepackage{tikz}
\usepackage{pgfplots}
\usepackage{caption}
\usepackage{subcaption}
\usepackage{wasysym}
\usepackage[margin=1in]{geometry} % for page margin
\usepackage{amsfonts,amsthm,amssymb}
\usepackage{hyperref}
\newtheorem{definition}{Definition}

\definecolor {myblue}{RGB}{92, 99, 143}
\definecolor {mymagenta}{RGB}{223, 67, 126}
\definecolor{sage}{RGB}{135, 174, 115}
\newcommand{\paragraphe}[1]{\vspace{5pt} \paragraph{\textbf{\textit{#1}}}}

\title{What Does it Mean for a Language Model to Preserve Privacy?}
\author{
    \textbf{Hannah Brown}$^1$, 
    \textbf{Katherine Lee}$^2$, 
    \textbf{Fatemehsadat Mireshghallah}$^3$\\
    \textbf{Reza Shokri}$^1$,
    \textbf{Florian Tram\`{e}r}$^4$\footnote{Authors appear in alphabetical order}\\
    $^1$National University of Singapore, 
    $^2$Cornell University \\
    $^3$University of California San Diego,
    $^4$Google\\
    \tt \{hsbrown, reza\}@comp.nus.edu.sg kate.lee168@gmail.com\\ 
    \tt fatemeh@ucsd.edu tramer@google.com
}

\begin{document}

\date{}

\maketitle

\begin{abstract}
	% In fact,  privacy considerations of natural language are as broad as that of real life. 
	Natural language reflects our private lives and identities, making its privacy concerns as broad as those of real life.
	Language models lack the ability to understand the context and sensitivity of text, and tend to memorize phrases present in their training sets.
	% The harm of memorization can vary depending on 
	An adversary can exploit this tendency to extract training data. Depending on the nature of the content and the context in which this data was collected, this could violate expectations of privacy. 
	%Humans delicately regulate the exchange of private information, depending on the \emph{context} of conversation and the underlying \emph{trust} between individuals and communities. 
	Thus, there is a growing interest in techniques for training language models that \emph{preserve privacy}. 
	In this paper, we discuss the mismatch between the narrow assumptions made by popular data protection techniques (data sanitization and differential privacy), and the broadness of natural language and of privacy as a social norm. 
	We argue that existing protection methods cannot guarantee a generic and meaningful notion of privacy for language models. 
	% We show that these  protection methods can preserve, at best, an limited form of privacy which does not account for nuances in human language, thus cannot claim to guarantee a generic and meaningful notion of privacy. 
	We conclude that language models should be trained on text data which was explicitly produced for public use.
\end{abstract}

\section{Introduction}

We use natural language to construct identities and communicate all our information in day-to-day life.
Humans naturally understand when sharing a sensitive piece of information is appropriate based on context. 
It may be fine to share the same piece of information with one specific person or group, and a complete violation of privacy to share in another context, or at another point in time.
Between humans, we trust that these implicit boundaries will be recognized and respected.
As we build technologies that collect, store, and process our natural language communication, it is important that these technologies do not violate human notions of privacy or make use of data in ways beyond what is needed for the utility of the technology~\cite{nissenbaum2009privacy, zuboff2019age}.

Language models (LMs) underlie much natural language technology we regularly interact with, from autocorrect to search engines and translation systems.
Over the past few years, LMs have grown in size and now utilize unprecedentedly large datasets of natural language making privacy risks in LMs a far reaching problem.
Prior work has already demonstrated that such models are prone to memorizing and regurgitating large portions of their training data~\cite{carlini2019secret,carlini2020extracting, lee2021deduplicating, henderson2018ethical, thakkar2021understanding}. 
Worse, they are especially likely to memorize atypical data points---which are more likely to represent privacy risks for the authors or subjects of these texts.

To address these privacy concerns, 
there is a growing body of literature that aims to create \emph{privacy-preserving} language models~\cite{mcmahan2017learning, anil2021large, li2021large, yu2021differentially, shi2021selective, hoory2021learning, ramaswamy2020training}.
While humans navigate the complexities of language and privacy by identifying appropriate contexts for sharing information, LMs are not currently designed to do this~\cite{curry_metoo_nodate,nobles_responses_2020,miner_chatbots_2020,kocaballi_responses_2020,miner_chatbots_2020,latitude_2019,hovy2021importance}.
Instead, the approach to preserving privacy in LMs has been to \textit{attempt} complete removal of private information from training data (data sanitization), or to design algorithms that do not memorize private data, such as algorithms that satisfy differential privacy (DP)~\cite{dwork2006calibrating, dwork2008differential}.

\textbf{Both methods make explicit and implicit assumptions about the structure of data to be protected, the nature of private information, and requirements for privacy, that do not hold for the majority of natural language data.}
Sanitization techniques assume that private information can be formally specified, easily recognized, and efficiently removed. 
In contrast, the semantic privacy guarantee offered by DP is that an adversary cannot distinguish whether any individual \emph{record} was used to train an LM, which implicitly assumes that these records are well defined and logically map to individual pieces of private information to be protected.

We argue that while these methods can provide some limited form of \textit{data protection} for specific types of text data, they \emph{cannot fully satisfy} the \textit{privacy} expectations that humans endow on the text they share.   Data sanitization is only able to recognize a vanishingly small portion of textual private information.
In turn, differential privacy can only provide meaningful protection guarantees for information that has clearly defined borders, thereby ignoring the reality that text is inherently a means of \textit{communication}, and that sensitive information is routinely written by or shared among groups of individuals, which blurs the borders of private information.  Instead, we argue that an appropriately named ``privacy-preserving'' LM should guarantee that a user's data cannot ever appear (or be inferable) outside the context they originally expected it to appear in (i.e., respect \emph{contextual integrity}~\cite{nissenbaum2009privacy} in the presence of inference attacks)---an ability that cannot be achieved without a deep understanding of the context in which the private information is produced, used, and shared.

Users' private data is being constantly used to train and fine-tune various services based on language models, which can obviously violate data privacy.  Instead, public sources of data (e.g., Web scrapes), seem to not pose privacy risks. 
Yet, public availability of language data should not be mistaken for data intended to be made public. 
Text may be shared by humans specifically to violate someone else's privacy (e.g., doxing), and even public social media posts are not always intended for an audience broader than one's acquaintances.
Even if this is not the case, applications of LMs could make data usable or searchable in new, unintended ways, or make it harder for the data to be modified or erased.
An understanding of context is necessary to judge whether it is appropriate to use a piece of data in training.

We further argue that individual users cannot give informed \emph{consent} for their data to be used in a LM or not. 
First, researchers are still working to quantifying the privacy risks of allowing one's data to be part of a LM training set.
Second, one user's private information is likely contained in the text of many other users. 
A single user would not be able to specify how all the text they have contributed is managed.  
We thus conclude that \textbf{data protection is not equivalent to privacy protection for natural language data} and to offer any \emph{meaningful guarantee of privacy}, LMs should be trained on data that was explicitly intended for fully public use, both at present and into the future.

\section{Background on Language Models} \label{sec:LM}

Language models (LMs) are essential components of state-of-the-art natural language processing pipelines, and refer to systems that are mainly trained on a large corpus of text for word sequence prediction tasks.  More precisely, a language model is optimized to learn the occurrence probability of tokens\footnote{a token is an instance of a sequence of characters that are grouped together as a useful semantic unit for processing -- it could be a character, a word or a sub-word.} in any sequence, based on the co-occurrence of tokens in the training data.  The ultimate objective is to find the relation between a token and its preceding or surrounding segments.  To this end, language models extract various statistics and correlations from sequences of words, at the level of sentences or paragraphs.  

The current trends of language modeling also shows that aggressive data collection and training enormous models are crucial for improving the performance of LMs. 
State of the art algorithms based on large neural networks enable effective extraction and encoding of a vast number of statistics about the training corpus, and have achieved unprecedented performance on a wide range of applications.  
The pervasive application of LMs and ever-larger datasets needed to train them pose serious privacy concerns.

\paragraphe{Applications of Language Models.}
There is a significant interest in the research community and industry to \textbf{apply LMs in any situation where humans use natural language} such as: assisting humans in various services, or facilitating communication.  
For example, LMs are being used in call centers, medical applications, mobile phones and personal computers and home assistants (such as Apple Siri, Amazon Alexa, Google Assistant, Microsoft Cortana, etc), email and message auto-complete services, document translation and search, writing companions (such as SmartCompose~\cite{dai-smartcompose-2019}, Codex and CoPilot for code completion~\cite{chen2021evaluating}), and many other situations where personal and sensitive data is created and used.  
The following is a short list of some common language model tasks, which are the foundation of many of LM applications: part of speech (POS) tagging and parsing~\cite{durrett-klein-2015-neural}, optical character recognition (OCR)~\cite{memon2020handwritten}, automatic speech recognition~\cite{park2019specaugment}, natural language generation~\cite{brown2020language}, sentiment analysis~\cite{yadav2020sentiment}, and natural language inference~\cite{bowman2015large}.  Applications based on these tasks process potentially private data at scale, such as user queries, sensitive documents, emails, and private conversations.

\paragraphe{Objectives and Types of Language Modeling.}
Language models are trained to construct sentences that resemble natural language. 
They do this by learning statistical measures to capture the local role of each word among its surrounding words and its global consistency within a longer sequence of words (e.g., the whole sentence, or the paragraph).  

One core feature of LMs is learning \textbf{embedding functions}: mappings from words (and phrases) to vectors in a high-dimensional space such that the closeness between two vectors reflects how close the meanings of the corresponding words (and phrases) are.  Embedding functions act as a proxy to encode the semantics of words and sentences in a language and are based on the particular sentences observed in a training corpus.  So, training reliable embedding models requires a significant amount of training data.  The embedding functions are then used as inputs for downstream NLP tasks.  Two other major state-of-the-art classes of neural language models also enable \textbf{generating and representing text}: generative LMs which focus on next-token prediction (for example, transformer-based models~\cite{vaswani2017attention}, such as GPT-3~\cite{brown2020language}) and masked LMs with the objective of filling in blanks in a sentence (for example,  BERT~\cite{devlin2018bert} and RoBERTa~\cite{liu2019roberta}).

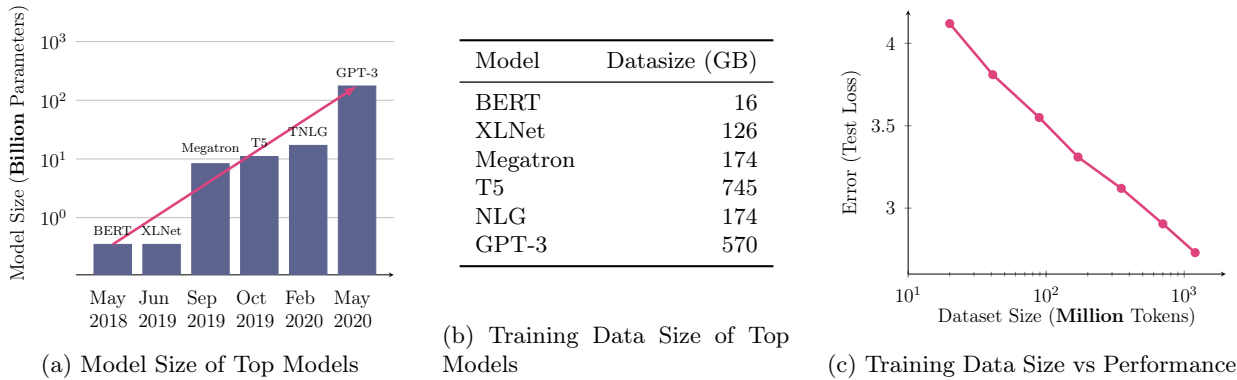
\begin{figure}
    \centering
    \begin{subfigure}[b]{0.33\textwidth}
        \centering
        \resizebox{\textwidth}{!}{
        	\begin{tikzpicture}
	\LARGE
    \begin{semilogyaxis}[
        %The axis design
        x tick label style={text width=1cm, font=\large},
        y axis line style={draw=none},
        y tick label style={font=\large},
        ylabel style={font=\large},
        tick style={draw=none},
        axis x line = bottom,
        ymajorgrids,
        ylabel={Model Size (\textbf{Billion} Parameters)},
        ymax=1e3,
        %A bar plot
        ybar,
        bar width=23pt,
        symbolic x coords={May 2018,Jun 2019,Sep 2019,Oct 2019,Feb 2020,May 2020},
        enlargelimits=0.15,
        % For the bar label (BERT etc.)
        nodes near coords,
        log origin=infty
    ]

    \addplot[color=myblue,fill,
        point meta=explicit symbolic,% For the bar label (BERT etc.)
        text= black] 
        coordinates {(May 2018,.35) [\footnotesize BERT]
                    (Jun 2019,.35) [\footnotesize XLNet]
                    (Sep 2019, 8.3)    [\footnotesize Megatron]
                    (Oct 2019,1.10e1)   [\footnotesize T5]
                    (Feb 2020,1.70e1)   [\footnotesize TNLG]
                    (May 2020,1.75e2) [\footnotesize GPT-3]};
    %The arrow		        
    \draw[mymagenta,-latex,ultra thick] (axis cs:May 2018,0.35) -- (axis cs:May 2020,1.75e2);
    
    \end{semilogyaxis}
    \end{tikzpicture}
        }
        \caption{Model Size of Top Models}
    \end{subfigure}
   	\hfill
   	\begin{subfigure}[b]{0.28\textwidth}
    	\centering
    	\raisebox{13ex}{
     	\resizebox{\textwidth}{!}{
    		
%\newcolumntype{O}{>{\RaggedLeft\arraybackslash}m{0.07\linewidth}} 
%\newcolumntype{D}{>{\arraybackslash}m{0.15\linewidth}} 
%\newcolumntype{R}{>{\arraybackslash}m{0.29\linewidth}} 
{\small
\begin{tabular}{lr}
  \toprule
  Model & Datasize (GB)  \\
  \midrule
  BERT & 16\\
  XLNet &  126\\
  Megatron & 174\\
  T5 &  745\\
  NLG & 174\\
  GPT-3 & 570\\
  \bottomrule
\end{tabular}
}
     	}}
    	\caption{Training Data Size of Top Models}
   	\end{subfigure}
   	\hfill
   	\begin{subfigure}[b]{0.33\textwidth}
    	\centering
    	\resizebox{\textwidth}{!}{
    		\begin{tikzpicture}
	\LARGE
    \pgfplotstableread{./data.csv}{\data} %Import the data from a tab separated file
    \begin{semilogxaxis}[ % Semilogsx creates a logarithmic x axis
    xlabel style={font=\large},
    ylabel style={font=\large},
    x tick label style={font=\large},
    y tick label style={font=\large},
    xlabel= {Dataset Size (\textbf{Million} Tokens)},
    ylabel= {Error (Test Loss)},
    %Only axis on the sides (instead of a box)
    axis x line = bottom,
    axis y line = left,
    ymin=2.6,
    ymax=4.2,
    xmin=1e1,
    xmax=2e3,
    ]
    \addplot[ultra thick, color=mymagenta,mark=*]  table [x={x}, y={y}] {\data};
    \end{semilogxaxis}
\end{tikzpicture}
    	}
    	\caption{Training Data Size vs Performance}
    	\label{fig:trends:last}
	\end{subfigure}
   
	\caption{Recent trends in model size and training data size of language models (a and b), and the impact of training set size on model performance (c).  \textbf{State-of-the-art language models require a significant amount of training data. The size of top models also increases by an order of magnitude every year}.  These factors significantly increase the privacy risks of language models.}
    
	\label{fig:trends}
\end{figure}

\paragraphe{Trends in Language Modeling.}
Algorithms for learning language models (notably transformer LMs) show an unprecedented performance on extremely large models with hundreds of billions of parameters trained on extremely large datasets~\cite{bender2021dangers, HAN2021, desai2020calibration,zhong-etal-2021-larger, kaplan2020scaling}. Figure~\ref{fig:trends} illustrates this trend.  What is very important to note is that \textbf{using large models, large datasets, and high amounts of compute time are all essential for achieving a high performance}~\cite{kaplan2020scaling}. Empirical results show that the error (test loss) of a transformer-based language model has a power-law relationship to its model size, dataset size, and the amount of compute used for training (see, for example, Figure~\ref{fig:trends:last}). Thus, an order of magnitude scale-up is needed to observe tangible improvements in model performance.

\section{Privacy Risks of Language Models for their Training Data} \label{sec:privacy_violations_lm}

Machine learning models learn by extracting generalizable patterns from their training dataset.  Yet, it has also been posited that memorizing some parts of the training data can be necessary to optimally generalize to long-tailed data distributions~\cite{feldman2020does}.  For example, nearest neighbor language models~\cite{khandelwal2019generalization} which retrieve samples directly from their training dataset are shown to outperform their conventional counterparts.  Data memorization can directly lead to leakage of private information from a model's training set, where behavior of the model on samples that were present in the training set becomes distinguishable from samples that were not.   Such leakage has been demonstrated in high-dimensional machine learning models~\cite{shokri2017membership}, and recent large LMs~\cite{carlini2020extracting}.  The trend appears to get worse as both the size of LMs and their training sets increase (Figure~\ref{fig:trends}). Below we discuss concrete examples of such privacy risks and their consequences.

\paragraphe{Membership inference.}

Membership inference attacks reveal whether or not a given data-point was used in training a given model~\cite{shokri2017membership}.  These attacks can be seen as privacy risk analysis tools~\cite{murakonda2020ml}, which help reveal how much the model has memorized the individual samples in its training set, and what the risk of individual users is~\cite{shokri2017membership, nasr2019comprehensive, memberinf2, memberinf3, memberinf4}.  An adversary who has no direct access to the model and its training, for example in the case of machine-learning-as-a-service, is able to identify the members of the training data by simply querying the model~\cite{shokri2017membership}. Membership inference attacks are alarmingly powerful against neural network models with large capacity, enabling them to identify atypical (thus sensitive) members of the training set~\cite{nasr2019comprehensive}.  The power of membership inference attacks have been demonstrated on natural language processing models such as NLP classifiers~\cite{shejwalkar2021membership} as well as released embeddings~\cite{song2020information, mahloujifar2021membership}. Such attacks could cause especially serious harm if they are mounted on clinical models, where membership in the training set could reveal a person's disease condition.

\paragraphe{Training data extraction.}
Training data extraction refers to the risk of partially extracting training samples by interacting with a trained language model~\cite{saleme2020, carlini2019secret, santiago-snapshot-2020, carlini2020extracting}.  An adversary can use use membership inference attacks as an oracle to generate sentence samples that have a high chance to be in the training set.  This attack is demonstrated on the GPT-2 (Generative Pre-trained Transformer) language model family, which consists of three generative models, with different sizes~\cite{carlini2020extracting}.  The attack can successfully recover a person's full name, address, and phone number from the largest GPT-2 variant (Table~\ref{tab:examples}).  The empirical results show that the larger the model is, the more training samples it memorizes: demonstrating once again \textit{the curse of high-dimensionality for data privacy}.  Mounting the same type of data extraction attack on BERT-based models trained on de-identified clinical notes shows that more than 4\% of generated sentences with a patient's name also contain one of their true medical conditions~\cite{lehman2021does}.

Algorithms behind inference attacks only improve over time.  Thus, current attack results under-estimate the privacy risks of large machine learning algorithms, notably language models.  Given the privacy risks of LMs, there is an increasing attempt towards designing \textit{privacy-preserving} language models, which can learn the overall distribution and structure of human language, yet do not memorize sensitive information.  This can help preserving some notions of privacy, and preventing the out-of-context exposure of training data to unauthorized users.

Existing techniques for building privacy-preserving language models fall into two broad classes: (1) \textit{data sanitization} techniques that find pieces of private information in text and remove these before any further processing, and (2) \textit{differentially private} training algorithms that mitigate the risks of memorization. Section~\ref{sec:critique} dives deeper into these approaches, and argues that neither is adequate for creating language models that properly preserve users' privacy.

\begin{table}[t!]
	\renewcommand{\arraystretch}{1.25}
\begin{tabularx}{\textwidth}{@{} c r r c X @{}}
\bf Formatted & \bf Owners & \bf In-group & \bf In-group sharing & \bf Examples\\
\toprule
\CIRCLE & 1 & 1 & - & Personal password file, secret key\\
\CIRCLE &  1 & >1 &  \CIRCLE & SSN, password, credit card sent to others\\
\CIRCLE & 1 & $\infty$ &  \LEFTcircle &  A developer posts their name, address, and phone number as contact information on Github. Their personal information is ``public'' on the Web, but in a well defined context. %This information was extracted from GPT2 by Carlini, et al.\ref{carlini2020extracting}.
\\
\CIRCLE & >100 & >100 &  \CIRCLE & A company credit card is shared with employees.\\
\midrule
\Circle & 1 & 1 & - & Personal search history\\
%\Circle & 1 & >1 & \CIRCLE & Alice's primary care doctor shares medical information relevant to her current visit with her attending doctor.  \\
% \Circle & 1 & >1 & \Circle  & Alice shares her medical information with her doctor, and with her friend Bob. Her doctor and Bob are prohibited from discussing this information.\\
\Circle & 1 & 2 & \CIRCLE & Bob suffers a mental health crisis and texts a support hotline. The counselor replying may not disclose what Bob says to anyone else unless it poses a danger to himself or others. \\
\Circle & 1 & 3 & \CIRCLE  & An employee at Enron~\cite{klimt2004enron} shares their wife's social security number (who is not part of the company) for the purpose of setting up insurance.\\
\Circle & 1-2 & >1 & \Circle  & Alice texts her friends Bob and Charlie about her divorce. Bob further texts Charlie about the matter (c.f. Figure~\ref{fig:removing_data})\\
\Circle & >100 & >100 & \CIRCLE & The Panama papers are discussed by 300 reporters for a year before being publicly released. \\
\bottomrule
\end{tabularx}  
\caption{
Examples of private information, and the contexts in which they might be shared.
A piece of private information is ``owned'' by one or more users (e.g., a credit card that belongs to one user vs. a company credit card that is shared by many). Private data can be shared within a group (the in-group) of variable size. Members of the in-group may be allowed or prohibited from further sharing or discussing the information with other members of the group.
Private data can be ``formatted'' such as a social security number (SSN), or a credit card number, or be referenced in arbitrary prose.}
\label{tab:examples}
\end{table}

\begin{figure}[hbt]
	\vspace{0.5in}
	\centering
	\begin{subfigure}[t]{0.23\textwidth}
		\centering
		\includegraphics[width=\textwidth]{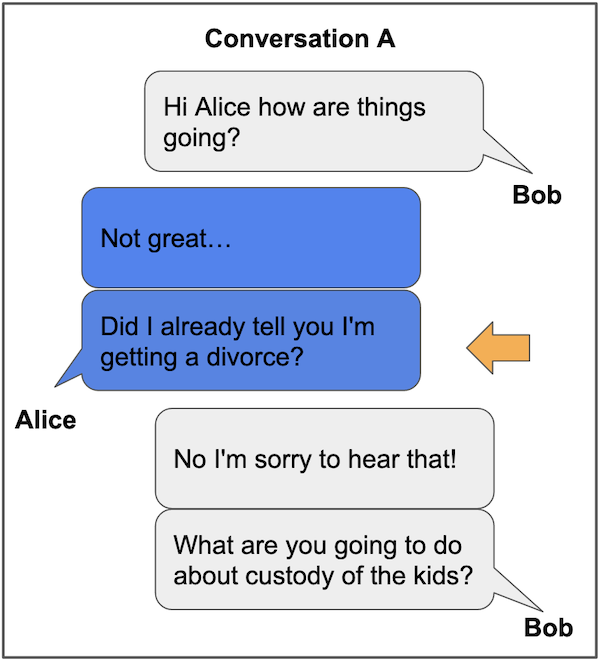}
		\caption{Original conversation}
		\label{fig:nothing_removed}
	\end{subfigure}
	%
	%\quad\quad
	%
	\begin{subfigure}[t]{0.23\textwidth}
		\centering
		\includegraphics[width=\textwidth]{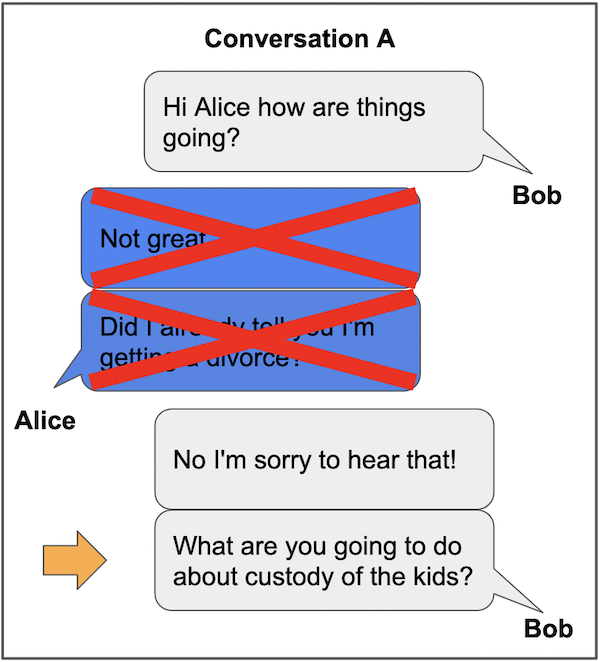}
		\caption{Alice's messages removed}
		\label{fig:message_removed}
	\end{subfigure}
	\begin{subfigure}[t]{0.475\textwidth}
		\centering
		\includegraphics[width=\textwidth]{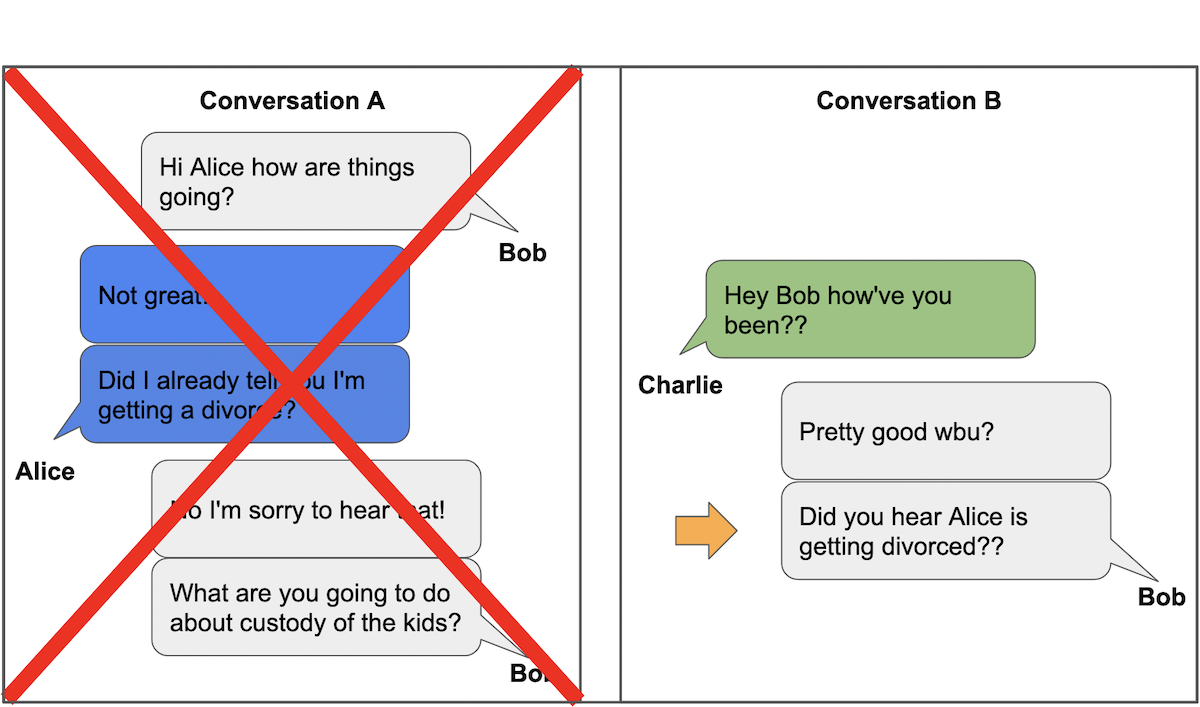}
		\caption{Alice's information is shared by Bob}
		\label{fig:conversation_removed}
	\end{subfigure}
	\caption{Illustration of the difficulties in removing private information from a dataset. Private information indicated by orange arrows: (a) The original conversation, where Alice shares her private information with Bob. (b) The conversation with all of Alice's messages removed. Bob's last message still includes her private information. (c) The whole original conversation is removed. Conversation B still contains Alice's private information though she is not in the conversation. }
	\label{fig:removing_data}
\end{figure} 

\section{What does preserving privacy in language modeling require?} \label{sec:privacy_requirements}

To claim a language model is privacy preserving, it must only reveal private information (aka ``secrets'') in the right contexts and to the right people.
While this goal is easy to state, the definition is comprised of three parts, each of which is challenging to determine: (1) in what contexts a secret can be shared without violating privacy (2) what information is contained in the secret, and (3) which people know the secret (the "in-group"). 

Far too often, the standard for data protection extends only to not revealing information that harms an individual. 
Inference attacks, such as those described in Section~\ref{sec:privacy_violations_lm}, show the possibility of information leakage in language models.
It is not enough to claim privacy is preserved because attacks are not able to extract information from a model.
These attacks improve over time, so while a model that current attacks can extract only a small amount of data from is at low risk for \emph{privacy violation}, this is insufficient for the claim that the model \emph{fully preserves privacy}.

In this section, we illustrate the wide variety of forms private information may take, and how only by understanding context and following privacy norms, can we construct language models that fully preserve privacy.
Finally, we discuss how humans approach decisions about when to reveal private information and draw parallels to language models and common privacy defenses. 
To motivate our main arguments, we provide illustrative examples of different types of personal information shared via natural language in Table~\ref{tab:examples}. These examples cover four axes of variation:

\begin{itemize}
 \item Some secrets (typically) follow a specific format (e.g., a credit card number), while others are embedded in prose.
 \item Secrets relate to (or are owned by) a single individual or multiple.
 \item Secrets are shared with a group (the ``in-group'') of one or more individuals.
 \item Individuals in the in-group may be allowed or prohibited from further sharing or discussing the secret among themselves either implicitly or explicitly (e.g., via legal restrictions).
\end{itemize}

\subsection{Secrets are contextual}\label{sec:context}

Respecting privacy requires being aware of the context in which information is shared~\cite{dourish2004privacy,nissenbaum2009privacy}. 
Instead of simply not ``memorizing'' private information, humans keep information private through complex judgments of appropriateness dependent on conversational and socio-cultural context. 
These judgments require information beyond the text of a conversation, making it impossible for an observer, human or computer, to make these same judgments absent this context.
Revealing a piece of information to some people may be fine, while it may be a slight violation of privacy to reveal it to a broader audience, and a more severe violation still to make it completely public. 
These perceptions of privacy are important when considering potentially private information in textual training data. 
\textbf{The scope users mean to share their data in must be considered when deciding whether or not to use it in training.}

\paragraphe{Privacy is not a binary variable.}
Information that is readily shared in one context may be private in another. 
The counselor texting Bob to help him cope with his mental health crisis (Table \ref{tab:examples}) may share details about his situation with other professionals or emergency services if there's reason to believe Bob poses a risk to himself or others, but is otherwise prohibited from sharing what Bob texts.
A specific identifying piece of information such as a phone number would be considered sensitive
if it belongs to a private individual, and benign if it belongs to a public entity such as a company.
More broadly, pieces of data can lie on a spectrum of privacy \emph{levels} with different restrictions and expectations, between the two extremes of fully public (e.g., Wikipedia) or fully private (e.g., someone's search history).

Training a language model for public use on data that was not intended for that level of publicity violates the original privacy expectations of that data. 
Nissenbaum's contextual integrity~\cite{nissenbaum2009privacy} provides a framework for disambiguating which contexts information can be shared under and with whom.
Under contextual integrity, there are five features (the data subject, sender, recipient, information type, and transmission principle) such that if any one is modified, 
the expectation of privacy could change.
In practice, this context could be indicated in the form of social cues and norms, or through regulations (such as Health Insurance Portability and Accountability Act (HIPAA) for medical data) or non-disclosure agreements for corporate information.
Under this framework, privacy is considered violated when information is shared outside an acceptable context, which also allows some concept of different degrees of privacy violation\cite{solove2006taxonomy}.

\paragraphe{Language models do not understand context.}
In practice, building machine learning systems that are sufficiently aware of context to appropriately judge the privacy of a piece of information is challenging.
Outside of privacy concerns, detecting implied context and reacting to it appropriately has been an active area of research for language models~\cite{weidinger2021ethical}. 
For example, work has been done to assess how appropriately chatbots respond to delicate situations, such as responding to sexual harassment~\cite{curry_metoo_nodate}, discussions of suicidal intent or other mental health issues~\cite{nobles_responses_2020,miner_chatbots_2020}, and mentions of violence or physical danger~\cite{kocaballi_responses_2020,miner_chatbots_2020}. 
The results were less than encouraging, with Nobles et al.~\cite{nobles_responses_2020} finding that the majority of the time, chatbots responded in inappropriate ways in these situations, ranging from simply saying "can you repeat that" to giving actively harmful information.
Another related line of work is context-aware, long-form, ethical  and persona-based, response generation~\cite{majumder-etal-2020-like,generation_2020,jiang2021delphi, li-etal-2016-persona}, where the chatbot or dialog system is supposed to hold a conversation with previous context taken into consideration~\cite{latitude_2019}.
This context could be the persona of the user, or previous conversations. 
Although this task has advanced in the past few years, the proposed models are plagued by the same challenges as the chatbots and the problem is far from solved~\cite{hovy2021importance}.

As another example, consider the case of Alice telling Bob about her divorce, as illustrated in Figure~\ref{fig:removing_data}.
Assuming we had a way to recognize that Alice getting a divorce is private information, we could remove Alice's message to Bob about her divorce.
However, this still leaves two more messages referencing the same sensitive information.
Bob's message to Charlie explicitly says that Alice is getting a divorce, and thus obviously refers to the same secret. Bob's reply to Alice asking about the custody of her kids is more subtle.
Understanding that this message is sensitive necessarily requires broader knowledge about the contexts in which asking about custody may occur, and more personalized knowledge about which context most likely applies to Alice.
The content of Bob's message can thus be considered just as sensitive as Alice's original message, yet automatically identifying the sensitivity of Bob's message may be much more challenging.
\subsection{Secrets are hard to identify}\label{sec:secrets_no_borders}

There are many ways of articulating the same point and additional phrasings can be added as language continues to evolve. 
Private information communicated through language is no different. 
This can make it difficult to identify whether a piece of text corresponds to private information.

\paragraphe{Form and Meaning: There are many ways to communicate any piece of information.}

Even private information that has an ascribed format (phone numbers, email addresses, credit card numbers, etc.) can appear in multiple forms. 
For example, numbers and symbols can be spelled out and content can be alluded to, eg: first initial last name at gmail dot com. 
Synonyms for words can be used, changing the appearance of text but not the meaning.
Anaphoric and cataphoric references 
\footnote{Using a word/phrase to refer to a named entity that is named earlier or later (respectively) in a conversation. In the case of anaphoric references, this entity may also have been defined in a prior interaction.} 
present a further challenge to recognizing private information.
If Bob were to instead say "Did you hear she's getting a divorce?" to refer to Alice in Figure~\ref{fig:removing_data}, that information would still be just as private and involve Alice just as much, but it is harder to automatically recognize.

Format-free pieces of private information, such as those referenced in Table~\ref{tab:examples} are even more difficult to identify and delineate.
Figure~\ref{fig:removing_data} shows how drawing a boundary around the text that references a given secret can be difficult. 
If Alice's divorce is sensitive information, then should mention of her custody battle be as well?
What about future conversations where Alice is referred to as ``single''?
Imbuing a language model with enough societal context and awareness to recognize these connections appears challenging.

\paragraphe{Repeated information can still be private information.}

Stating the same private information time and again does not make it less private.
One example is a company credit card number. 
This number might be shared again and again within the company, but it remains private to people outside the company.
Consider also the case of the Panama papers (Table~\ref{tab:examples}) which contained leaked legal and financial documents detailing how shell companies were created for illegal ends, like tax evasion.
Even though 300 journalists exchanged conversation about the Panama papers over the course of a year, the topic of these conversations is no less sensitive. 
If a language model had been trained on the journalists' emails, the topic of the investigation (or individual names of suspects or sources) could have been memorized and leaked.
De-duplicating the training dataset would not necessarily reduce the likelihood of this information from being learned, as the examples in the dataset are not necessarily near or exact duplicates of each other, but just happen to reference the same broad sensitive topic.
Respecting privacy in this case again requires both the ability to recognize the private information and to gather all training data items that refer to this private information.

\paragraphe{Language evolves, and so does private information.}
Changes in language or in social norms can shift the way in which people talk about secrets (or whether something is considered secret or not).
For example, the word `queer' was reclaimed by some members of the LGBTQ community beginning in the 1990s as part of the gay rights movement. A system that aims to automatically detect sensitive pieces of text would thus have to be aware of such shifts in linguistic meanings.
Yet, changes in language can be swift, in particular on social media~\cite{sayers2014languagechangemodel,eisenstein2014diffusion} where movements, such as \#MeToo or Black Lives Matter, can quickly and radically shift the meaning of words and phrases.
Additionally, to evade automated censorship and demonetization methods that target specific keywords and phrases, specific topics are routinely re-represented with new words and phrases~\cite{kim_trkic_2021,hiruncharoenvate_algorithmically_2015,tao_who_2021}.

Beyond the evolution of language, secrets can also evolve.
For example, while much of the content of the Panama papers investigation was highly sensitive and confidential while the investigation was ongoing, the findings were then made public. At the same time, the identities of the journalists' sources should remain secret essentially forever.

While languages and secrets naturally evolve, language models are typically trained once on a static dataset. Over time, these datasets, and thus the language models trained on them, become less useful for understanding current language. In Section~\ref{sec:critique}, we further explore how the use static datasets can present a challenge for privacy enhancing techniques such as data sanitization (Section~\ref{sec:sanitization}).
\subsection{In-groups are hard to identify}\label{sec:ingroups_no_borders}

Just like finding the borders of a secret is ill-posed, identifying the group of users who are privy to a secret (the in-group) is equally challenging. 
Indeed, individual text fragments can contain information pertaining to many individuals or organizations at once. 
The decision of whether to share the secret with a given individual varies from secret to secret, thus the in-group for each secret in different contexts is different. 
Additionally, just like the secret itself, the in-group can change and grow as relationships continue to evolve in the real world. 
Thus, even setting a reasonable bound on the \emph{size} of the in-group for each secret can be difficult. As we discuss in Section~\ref{sec:dp}, the lack of such a bound poses a particular challenge for articulating meaningful guarantees with differential privacy.

\paragraphe{Secrets can involve or be shared among many people.}
Natural language is \emph{meant to be shared}. 
We use language to articulate and communicate our thoughts and our observations. At times, these thoughts and observations can also be about other people.
Yet, many approaches to data privacy---in particular differential privacy---implicitly or explicitly assume that a user's private information does not transcend the user's own data (i.e., the user can protect their privacy simply by not sharing their own data). 
This assumption can be clearly violated in a variety of natural ways in which humans exchange textual information.

Consider the example described in Table~\ref{tab:examples} of an employee of a company who sends their wife's SSN to another employee.  We found an example of such an instance in an email from the Enron corpus~\cite{klimt2004enron}.
While the employee's wife might ``own'' her SSN, it now appears in the corpus of text written by the employee.
Typically, nothing prevents one user from sharing another person's private information (such sharing could be legally prohibited, or violate social trust, but these consequences do not mean sharing cannot occur). Thus it can be difficult to define a sole ``owner'' of a piece of private information.

What we say is often influenced by what others around us have said, which makes drawing dividing lines for privacy, much harder.
For example, social media whisper networks, like those discussed in~\cite{haire_media_men_2019} are almost exclusively devoted to sharing private information about people not in the network. 
In this case, a person outside of the network would still have their private information shared, and to complicate things, the collective information about this person could come from hundreds of different people's conversation data.
Another high-profile example is in the shadow profiles Facebook created of individuals who did not have Facebook accounts.
Without any personally volunteered data, Facebook was able to classify enough data to attribute it to an individual.
In Web-scraped datasets~\cite{radford2019language,raffel2019exploring} that are commonly used in training large language models~\cite{Dodge2021DocumentingTE} it is typically not possible to unambiguously map individual pieces of information to specific ``owners''.

\paragraphe{In-groups have no clear upper-bound.}
For any individual secret, we could attempt to identify the in-group of people who know the secret.
Given such knowledge, we could attempt to remove all mentions of the secret from the entire group. Alternatively, it could be tempting to provide privacy guarantees that are (inversely) proportional to the size of this group (e.g., as in differential privacy), following the intuition that information that has been shared many times is less sensitive than information that has been shared more rarely.

Yet this intuition fails to hold in regard to some of the examples listed in Table~\ref{tab:examples}, and there is no one number $k$ where a piece of information shared with $\geq k$ users can reasonably be assumed to be ``non-sensitive''.
One individual might share their closest secrets with a handful of friends or family members in a group chat. Others may share the same topics to a broader audience in a support group forum
or on their (private) social media page.
Companies and governments routinely share sensitive information with hundreds or thousands of employees. And more than 300 journalists communicated in secret for over a year before disclosing their findings in the Panama papers~\cite{panama}.
All this information is definitely private, but within specific contexts is allowed to be shared with a potentially large group of individuals.

\subsection{Human notions of privacy}\label{sec:ideal_privacy}

In contrast with common ML privacy preserving mechanisms which focus on preventing models from memorizing private information, humans very clearly memorize sensitive information that they learn.
Unlike LMs, we use learned conversational rules to gauge how appropriate or polite something is to share in a given context.
One of the simplest proposed sets of rules explaining how we speak---Grice's Maxims---are a set of four rules (together comprising the Cooperative Principle of Conversation) that describe ``normal'' conversation~\cite{grice_logic_1975}. 
Of these maxims, the ones we use to keep private information to ourselves are ``quantity'' (say exactly the amount appropriate in a given context) and ``relevance'' (say only what is relevant to the current context). 
These maxims are easy to state but \textit{heavily} context dependent, making 
them difficult to operationalize for technology.

Other conversational frameworks, like politeness theory or relevance theory~\cite{brown1987politeness,sperber1986relevance} also rely heavily on context, making their application to NLP systems challenging.
At a minimum, these frameworks require prior knowledge about the people involved in the conversation, the socio-cultural context, and past conversations---sometimes with people not involved in the current interaction, who may not have contributed themselves to the same dataset. 
Given only text data, and none of this further information, it is often impossible to gather all the context necessary to judge if saying something will violate someone’s privacy.
Furthering the idea that people memorize and use other methods to preserve privacy, previous work in psychology has shown that we are most likely to remember information that is either very in line with what we have seen before or very \emph{different} from what we've seen before~\cite{greve_knowledge_2019}. 
For example, when told a piece of surprising information that we know is supposed to be kept secret, we are likely to remember the information, but choose to not share it. 

In summary, humans respect privacy in natural language not by failing to memorize secrets, but by forming a judgment on whether any given piece of information is appropriate, or not, to share with a given party in a given context (unless they share it by mistake, or, by malice, intentionally).
Applying a similar approach to language models would require an intrinsic understanding of language and social contexts that goes beyond the capabilities of existing methods, as described in the next section.

\section{A Critical Analysis of Privacy Technologies for Language Models} \label{sec:critique}

Natural language processing algorithms that aim to respect privacy either remove private information from the  data (through text \textit{sanitization}~\cite{aura2006scanning, lison2021anonymisation, dernoncourt2017identification, norgeot2020protected}), or design learning algorithms that mitigate the risks of information leakage by not memorizing private information (through \textit{differentially private} learning~\cite{dwork2006calibrating, chaudhuri2011differentially, abadi2016deep, mcmahan2017learning}). 

In this section, we evaluate the \textit{claims} of these protection methods about preserving privacy, in the context of language data.  Our approach is to lay out the assumptions that data sanitization and differential privacy (DP) make (either implicitly or explicitly). Then, we  discuss how awareness of context, difficulty determining the borders of a secret and attributing it to individuals, and other  privacy nuances (as extensively discussed in Section~\ref{sec:privacy_requirements}), can invalidate these assumptions.  We discuss the kinds of privacy violations that each method would or would not protect against, and highlight that, given any specific definition for data, \textbf{data protection is not equivalent to privacy protection}.  They do overlap in many cases where a unit of data contains all the private information about an individual.  So, by removing it or not memorizing it (i.e., protecting it from being inferred), we protect the individual's privacy. However, in general, privacy is much broader than data protection, and this is notably the case in natural language.

\subsection{Data sanitization} \label{sec:sanitization}

Data sanitization claims to preserve privacy by removing private information.  The critical assumptions are that it is possible to \textit{formally specify private information}, and to \textit{design efficient algorithms to identify and remove private information according to the provided specifications}.  We evaluate how realistic these assumptions are, and question if data sanitization can preserve privacy in any meaningful way.

Based on the foundations of privacy in Section~\ref{sec:privacy_requirements}, we argue that private information expressed in text is difficult to specify and identify, and its removal (according to a given specification) is insufficient to preserve privacy in many situations.  Text data can be written in many forms, and the borders of private information are indeterminate.  This significantly narrows the application of data sanitization to limited cases where the secret is written according to a context-independent template (e.g., phone number written as consecutive digits).

\paragraphe{Sanitization is insufficient because private information is context dependent, not identifiable, and not discrete.}
Most data sanitization methods are algorithms that use parsers and classification models to tag each word in an input text either based on defined patterns or already tagged data (where sensitive words are manually identified).  These techniques work best for identifying well-formatted private information, such as social security numbers, and specific forms of medical note datasets~\cite{johnson2016mimic, dernoncourt2017identification, lison2021anonymisation, tourille2018evaluation}.  However, as we discuss in Section~\ref{sec:secrets_no_borders}, even well-defined information can be written in many formats or alluded to indirectly.  For example, identifying the social security number ``\textit{the first 2 digits are two two, and the remaining ones are three ...}'' is much more challenging than identifying ``$223\cdots$''.  So, even in cases where specifying private information is possible, their reliable identification might be very hard. 

Further, identifying and removing non-specific private information, such as the case of Alice's divorce and custody battle, or the entire discussion around the Panama papers, is significantly more challenging (if not impossible) for data sanitization schemes (which are based on classification models).  In general, secrets have no borders, and identifying the scope of relevant information is beyond the capability of taggers and parsers.  Besides, understanding sensitivity requires inferring the context, which is a very hard task for algorithms.  First of all, there is no formal way to define context, and supervised machine learning models are nonrigorous, empirically inaccurate, and non-explainable methods to classify sensitive information.  Secondly, the context related to a piece of text might not be present within the text, which makes understanding the context impossible even for humans.  Third, since taggers and parsers require defining ahead of time what the ``sensitive'' categories are, this limits what information might be related to other sensitive information.  Knowing that Alice's custody battle is sensitive requires understanding that there would be no battle if there were no divorce and requires cultural context (Section~\ref{sec:context}) that is beyond (current) algorithms.  Fourth, the context can change after data redaction, which consequently can change the sensitivity of text.  So, any claim for data privacy based on sanitization is always outdated. 

Changing the context of a piece of information can increase or decrease expectations to privacy. Bob may have a relatively small expectation of privacy when he makes a public social media post, but very high expectations when texting a crisis counselor. In this context, the act of sharing any data at all\footnote{The prominent text helpline, the Crisis Text Line, recently admitted to doing exactly this for the purposes of helping a for-profit company train language models to improve customer service~\cite{levine2022ctl,porcaro2022ctl}.} would be considered a privacy violation. Data sanitization completely ignores this, as it assumes information to be discrete and treats privacy as a binary variable.  This problem resembles the numerous failed attempts for anonymizing high-dimensional data by removing certain attributes~\cite{narayanan2008robust,garfinkel2019understanding}.  In the context of language data (with enormous number of dimensions), there is always a possibility of inferring sensitive information even if many pieces of text are redacted.  This means that it is possible that either we fail to achieve an acceptable level of privacy through sanitization, or a hypothetically privacy-preserving data sanitization might result in removing almost all the text, rendering it useless: ``sanitized data isn't''.\footnote{C.f., ``Anonymized data isn't'' -- Cynthia Dwork}

\paragraphe{Data sanitization is useful in very limited cases.}
We argue that it is not possible to claim privacy using data sanitization algorithms: there is not a specification that would allow private information to be redacted from free-form text data because private text data is not easily identifiable and requires additional context to determine if the information should be redacted.  However, data sanitization is a useful obfuscation method in the cases where pieces of context-independent, well-defined, static private information are to be removed from a text dataset. 

Data sanitization is currently widely adopted across industries as a data pre-processing step for removing Personally Identifiable Information (PII) or protected health information (PHI) by companies such as Microsoft, Paypal and Mastercard~\cite{balzer2020obfuscating, balzer2021protecting, austin2019self, donovan2021management, williams2021systems, williams2021systems, gkoulalas2021utility} and numerous start-ups (SkyFlow, Ground Labs, PII tools, MailTumble, etc.).  Data sanitization can remove some specified information, and can help to reduce the privacy risks to some (unknown) extent.  However, it cannot claim that it preserves privacy of individuals, as it has no formal definition for privacy which remains meaningful in the context of language data.

\subsection{Differential Privacy} \label{sec:dp}

Differential privacy (DP) is a data protection measure designed to assure users that contributing their data to a dataset will not reveal much additional information about the user, when the result of a DP algorithm trained on the dataset is released.  Put another way, the data protection guarantee offered by DP is that an adversary cannot easily distinguish whether any individual \emph{record} was used in the computation: 

\begin{definition}\label{def:dp}
	\textit{$\epsilon$-Differential Privacy}~\cite{dwork2006calibrating}. For a privacy loss parameter $\epsilon\geq 0$, a training algorithm $A$ satisfies $\epsilon$-DP if and only if for any pair of training datasets $D$ and $D'$ that differ in only one record, and any set of output models $S$: $\Pr[A(D) \in S] \leq e^\epsilon \Pr[A(D') \in S]$. 
\end{definition}

While many applications benefit from this protection, we argue that language data cannot be partitioned to ensure that algorithms trained with DP meet the standard of privacy we put forth in Section~\ref{sec:privacy_requirements}: \emph{to only emit private information to appropriate people in appropriate contexts.}  This is because sensitive language data, as we have seen, cannot necessarily be attributed to one individual or group, whether or not their data is included in the dataset.  Thus, while applications of some DP algorithms likely alleviate risks to privacy, they alone are insufficient for \emph{guaranteeing} the absence of privacy violations in language models.

\paragraphe{Differential privacy requires a unified definition for secret boundaries, which is very hard if not impossible to achieve for language data.}

The data protection guarantees of DP hold for any dataset $D$, and any content of the sensitive record.  Thus, compared with data sanitization approaches, DP sidesteps the issue of determining the context or content of private data by providing a worst-case guarantee that applies to any data record.  This enables applying DP algorithms in any setting where privacy is considered protected as long as each data record is protected.  

However, the main issue with applying DP to language data arises in how we define the \emph{boundaries} of private information. That is, how should we define what constitutes a data ``record'' in Definition~\ref{def:dp}.  Prior work has considered various granularities, from individual tokens or words, to sentences and documents, or all of a user's data~\cite{levy2021learning,mcmahan2017learning}.

Identifying data records with individual words or sentences makes sense from a machine learning perspective, since training batches are often split at such a granularity.  But the corresponding privacy guarantees are mostly inadequate since the removal of any individual word or sentence from the training data is insufficient to hide most types of private information (except maybe a password or SSN that falls inside a single data record).  It is thus much more appropriate to define DP with respect to all of a user's data.  Indeed, ``user-level" DP is the way in which the original DP definition is intended to be interpreted~\cite{dwork2006calibrating}.  In the context of language data, a user-level DP guarantee says that the trained model will be insensitive to the addition or removal of \textit{all the data written by any individual user}.  Yet, if we consider the examples in Table~\ref{tab:examples}, it is clear that many types of private information cannot be erased from a dataset by the removal of a single user (even after assuming that a ``user'' in the system/network is associated with a unique ``individual'').  Indeed, text is a means of \textit{communicating} information with others.  Thus, removing all of a user's messages is not sufficient to remove the private information from the training set, since others might reference the same information.\footnote{One could expand ``user data'' to encompass all \emph{conversations} that a user has participated in (including all replies they received), as in Figure~\ref{fig:conversation_removed}. First, satisfying such a level of DP is technically more challenging in decentralized settings (e.g., in Federated Learning~\cite{mcmahan2017learning}) since a data record now spans multiple participants (network users).  Second, such an increased granularity remains insufficient to protect knowledge of Alice's divorce (Table~\ref{tab:examples}) if this secret is further referenced in other conversations (such as between Bob and Charlie in Figure~\ref{fig:conversation_removed}.}  

\paragraphe{Protecting a specific unit of data is not the same as protecting privacy.}

The issue we are highlighting above is that private information can span across data provided by multiple individuals.  It is important to note, however, that the formal guarantees of DP hold regardless of such relationships in the dataset~$D$.  What is questioned here is what these data protection guarantees mean, semantically, for the protection of users' privacy.  Differential privacy can protect privacy to the extent that withholding one user's data from the dataset can.  Thus, it is useful for specific types of structured data, for example, when each individual contributes a record that contains sensitive attributes about them (e.g., whether they have been diagnosed with a particular disease). Or alternatively, when a user's secrets are indeed restricted solely to text written by that user (e.g., an individual's search history). These protections, however, cannot satisfy the full privacy expectations we discussed in Section~\ref{sec:privacy_requirements} regarding natural language data, where private information is not bounded by data records (and can even be about individuals who do not contribute any data), and collections of snippets of text that cover different pieces of private information might overlap.  So, withholding any specific unit of data from the dataset cannot \textit{guarantee} protection of privacy. 

\paragraphe{The need for privacy does not diminish with in-group size.}

The protection guarantees of DP for groups of users diminish exponentially as the size of the group increases ($k\epsilon$-DP for groups of size $k$).  However, in practice, the fact that some information is shared among more individuals does not necessarily make it less sensitive.  The sensitivity depends on the context and the reasons why the data provided by $k$ individuals contains the same private information.  Moreover, appropriately bounding the size $k$ of a group that is privy to a secret is also hard. For example, a community of individuals might share secrets at the level of the whole community.  In this case, DP does not provide any strong guarantees for protecting such secrets.

\paragraphe{On privacy guarantees and promises.}

Ideally, we would like to achieve ``secret-level'' differential privacy, i.e., the algorithm is insensitive to addition or removal of any piece of private information (e.g., Alice's divorce, or a company's secret).  But satisfying such a definition would require precisely understanding the context and boundaries of secrets, which is exactly a difficulty that DP aims to avoid.  

In a typical instantiation of user-level DP, the privacy guarantees provided by the training algorithm are hard to match to the above ideal. 
While some pieces of information enjoy strong formal protection guarantees from being definitely contained in one user's data (e.g., text in a user's personal search history), others are only protected at an exponentially small level (e.g., sensitive information shared among a large group).

This does not mean that information leakage is unbounded. In practice, the provable guarantees offered by DP algorithms are often estimated to be rather loose (i.e., the true leakage is less than what we can mathematically compute)~\cite{nasr2021adversary, ramaswamy2020training, chourasia2021differential}.  Yet, the main premise of DP is precisely that it provides provable guarantees, compared to the ad-hoc heuristic guarantees of many other privacy preserving techniques. These strong guarantees have at times been interpreted as a ``promise'' to users~\cite{dwork2011promise}, that their secrets will be protected regardless of their decision to share their data. As we have seen however, in the context of language data, this promise loses most of its meaning. We could then ask, whether the formal underpinnings of DP necessarily make it the privacy notion of choice for training LMs, or whether other approaches could provide for more semantically meaningful (albeit possibly only heuristic) forms of privacy protection.
 
\section{Summary and Discussion}\label{sec:discussion}

Underlying all the challenges of training language models that understand and respect privacy is the complexity of human privacy norms. The vast literature attempting to define privacy and provide frameworks for assessing and understanding it demonstrates the nuance required to disambiguate between what might be similar scenarios. Private information can take many forms, continuously change, and be shared by and among groups whose members fluctuate according to changes in human relationships.
In summary, \textbf{the boundaries of what data should be acceptable to use for a so-called ``privacy preserving'' language model are inherently fuzzy and context dependent}.

These challenges limit the applicability of existing techniques like data sanitization and differential privacy (Section~\ref{sec:critique}). Yet, \textbf{these privacy-enhancing techniques are often presented as providing users with certain \textit{guarantees} of privacy, which are \textit{not} meaningful enough given the assumptions they make about what constitutes private information}. It is true that, from the perspective of an individual user, the application of any obfuscation technique can only benefit privacy, compared to not applying them (e.g, training a model with DP is better than training the same model without DP). Yet, when applying privacy-preserving techniques to the collection of \textit{new} forms of data (as for training LMs on all types of text produced about every aspect of our lives), we need more realistic and rigorous privacy guarantees. %a more apt approach is simply to \textit{not} collect users' data at all.

What alternatives do we have? One might argue that models trained solely on publicly available data, such as text scraped from the Web, alleviate privacy concerns. And indeed, this is the approach taken by many recent large LMs~\cite{brown2020language, gpt-neo, raffel2019exploring, gopher}. 
Yet, publicly accessible does not mean public-intended: publicly shared data typically comes with an intended context of use, which language models could violate by memorizing data~\cite{carlini2020extracting}. Furthermore, the lack of public discourse and understanding around what happens with collected text data makes informed consent difficult to collect. Ideally, we want LMs to be trained solely on data that is intended (or allowed) for public dissemination. In addition, such LMs could be further fine-tuned or personalized locally on a user's non-public data, only if the model is going to be used by the same user. 
But disentangling data that is intended for public use, and obtaining appropriate user consent for its use remains challenging. We discuss these issues in more detail below.

\paragraphe{Publicly accessible data is not public-intended.}

Data that is publicly accessible (e.g., on the Web) is not necessarily intended for unfettered public dissemination, and its use in LMs could still pose privacy risks. For example, publicly available data might not be released by the data subject, such as leaked or subpoenaed email datasets~\cite{klimt2004enron, goss2020temporal}, copy/pasting conversations to distribute, or doxing an individual. Posts on social media can also sometimes be made public inadvertently~\cite{marwick_networked_2014, trepte_social_2021}. Furthermore, online text can be deleted or modified. A language model trained on earlier versions of such data would thus inadvertently serve as a data archive. 
Finally, models trained on Web data might also surface new unintended ways for this public data to be searchable. The example given in Table~\ref{tab:examples} where an individual posted their contact information on their Github is an actual example of training data extracted from a LM~\cite{carlini2020extracting}.

\paragraphe{Can users provide informed consent? Mostly not.}

Suppose that we asked users to opt-in to having parts or all of their data used out of context to train a language model.
For example, one mobile chat client might tell users that it will deploy privacy-preserving LM training on their chat messages, and users and their friends can decide to use the service, or not. Moreover, users might even have the option of flagging individual messages as acceptable for use in training or not. We argue that even if such a consent mechanism were to exist, it would be challenging for users to reach an informed decision about the consequences of their actions.

To start, even experts on ML privacy currently only have a partial understanding of the risks of data memorization and extraction (Section~\ref{sec:privacy_violations_lm}), and about how well various defense mechanisms perform. As we argue in Section~\ref{sec:dp}, even principled approaches such as differential privacy cannot provide privacy guarantees that are directly interoperable with privacy \textit{expectations} users might have for their text data. Moreover, individual users cannot properly consent to providing their sensitive information, since they are often not the only person holding that information. As we have illustrated in Section~\ref{sec:dp} and Table~\ref{tab:examples}, sensitive information is routinely shared among many users, all of which would have to jointly consent to release or withhold that piece of data. Put differently, the responsibility to share or hide private information always lies with the entire group that has knowledge of the information. Without understanding how their data will be stored, processed, and disseminated, people are unable to give informed consent.

\paragraphe{Private personalization.}

One approach that we view as a promising middle ground, and worthy of further exploration, is the development of LMs that are trained exclusively on data that is explicitly intended for public use, and further fine-tuned (or personalized) on users' local (private) data. As long as the model is only used in the local context of the user, the main privacy risks to the user would be alleviated.

\section{Conclusions}

Our entire life is mediated through language, much of which is monitored and processed by technology. \textbf{No discussion of privacy is complete without a deep analysis of how language data is handled.}
In this paper, we call for a rigorous understanding of privacy expectations, and for meaningful guarantees of privacy, in the context of language data. We highlight that data protection (with all its limitations) is not equivalent to privacy, existing so-called privacy-preserving methods do not provide reliable assurance about privacy, and users are not in a position to give consent for their data to be used for arbitrary computations. We argue that the only truly privacy preserving solution is to rely exclusively on data that is intended to be public.

\section{Acknowledgments}

The authors would like to thank David Mimno, Nicholas Carlini, Helen Nissenbaum, Vitaly Shmatikov, Noah Fiedel, Greg Yauney, Maria Antoniak, Federica Bologna, and Martin Strobel for discussions about different parts of this paper. 

\bibliography{references.bib}

\end{document}